\documentclass[conference]{IEEEtran}
\IEEEoverridecommandlockouts

\usepackage{cite}
\usepackage{amsmath,amssymb,amsfonts}

\usepackage{graphicx}
\usepackage{textcomp}
\usepackage{xcolor}

\usepackage{algorithm}
\usepackage{algorithmic}
\usepackage{multirow}
\usepackage{booktabs}
\usepackage{graphicx}
\usepackage{subcaption}
\usepackage{cleveref}
\usepackage{xspace}
\usepackage{authblk}

\begin{document}
\def \sysname{FLiP\xspace}

%\title{\sysname: Privacy-Preserving Federated Learning based on the Principle of Least Privileg}
\title{Privacy-Preserving Federated Learning via Dataset Distillation}

\author[1]{ShiMao Xu}
\author[1]{Xiaopeng Ke}
\author[1]{Shucheng Li}
\author[1]{Xing Su}
\author[1]{Hao Wu %\thanks{Corresponding author: email@mail.com}
}
\author[1]{Sheng Zhong}
\author[1]{Fengyuan Xu}

\affil[1]{Nanjing University, Nanjing , China}

\maketitle

\begin{abstract}
Federated Learning (FL) allows users to share knowledge instead of raw data to train a model with high accuracy. Unfortunately, during the training, users lose control over the knowledge shared, which causes serious data privacy issues. We hold that users are only willing and need to share the essential knowledge to the training task to obtain the FL model with high accuracy. However, existing efforts cannot help users minimize the shared knowledge according to the user intention in the FL training procedure. This work proposes FLiP, which aims to bring the principle of least privilege (PoLP) to FL training. The key design of FLiP is applying elaborate information reduction on the training data through a local-global dataset distillation design.
We measure the privacy performance through attribute inference and membership inference attacks. Extensive experiments show that FLiP strikes a good balance between model accuracy and privacy protection.
\end{abstract}


\section{Introduction}
\label{sec:intro}

\begin{figure*}[t]
     \centering
     \begin{subfigure}[b]{0.49\textwidth}
         \centering
         \includegraphics[width=\textwidth]{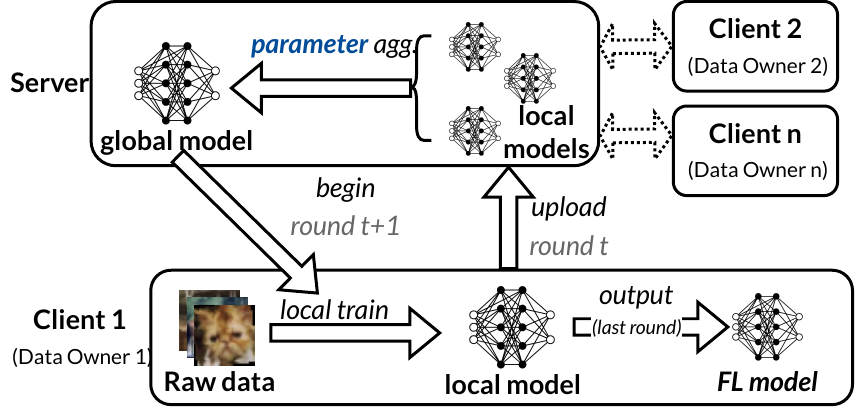}
         \caption{The vanilla FL paradigm.  }
         \label{fig:fl_model}
     \end{subfigure}
     \hfill
     \begin{subfigure}[b]{0.49\textwidth}
         \centering
         \includegraphics[width=\textwidth]{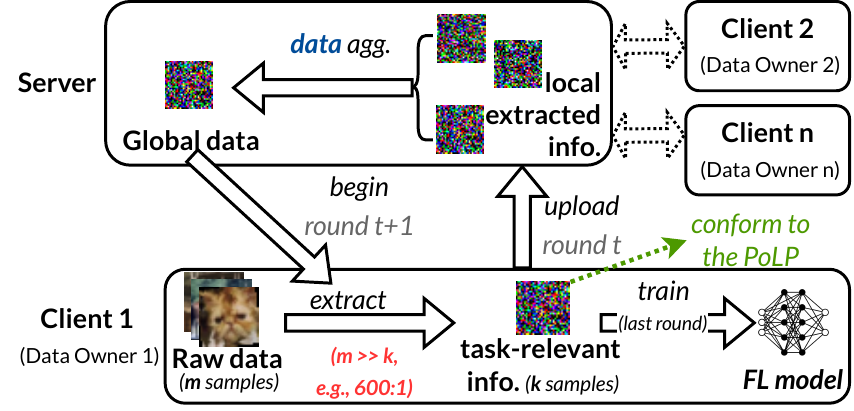}
         \caption{The FL paradigm of \sysname.  }
         \label{fig:fl_data}
     \end{subfigure}
        \caption{The training procedures of vanilla FL and \sysname. In both paradigms, no training is needed on the central server. \textbf{The biggest difference between our \sysname and the vanilla FL is the carrier of information aggregation, i.e., \sysname performs distilled data aggregation, and the vanilla FL  performs parameter aggregation.} Compared to the vanilla method, the amount of shared information during the training in \sysname is controllable.}
        \label{fig:fl_scenarios}
       \vspace{-10pt}
       % \Description{The training procedures of vanilla FL and \sysname.}
\end{figure*}

Federated learning (FL)~\cite{yang2019federated} is a deep learning (DL) training paradigm, which aims to utilize the data existing in the form of isolated islands to train DL models (Fig.~\ref{fig:fl_model}).
During the training procedure, data owners (a.k.a. clients) do not share their raw data with anyone, but instead, share some information obtained from the raw data in the form of model parameters.
% % However, there is a risk of data privacy leakage during FL training~\cite{lyu2022privacy}.
Many solutions are proposed to protect the privacy in FL context.
However, they are yet unable to strike a balance between performance and privacy protection in real-world scenarios.
This is because they do not fully consider the training task itself when performing protection. 
Specifically, considering that users are likely only interested in obtaining a high-quality model for the target task, sharing any information unrelated to the training task during the training process could potentially lead to privacy leakage, e.g., secondary attributes inference~\cite{lyu2021novel}, practical attribute reconstruction Attack~\cite{chen2022practical}.

\begin{figure}[t]
     \centering
    \includegraphics[width=\linewidth]{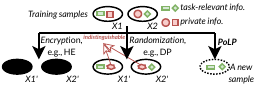}
    \caption{Comparison of PoLP and existing privacy protection solutions. Ideally, the new sample generated by PoLP only contains task-relevant information.}
    \label{fig:polp}
    % \Description{Comparison of PoLP and existing privacy protection solutions. Ideally, the new sample generated by PoLP only contains task-relevant information.}
\end{figure}

Given the subjective nature of privacy protection, we hold that an ideal solution should fully consider the user's training goals and only share essential information related to the training task during the training.
That's the core idea, what we call as principle of least privilege (PoLP).
According to the PoLP, as shown in Fig.~\ref{fig:polp}, clients should control only the essential training task-relevant information from the raw data that can be shared among participants. 
At first glance, it is paradoxical to determine which part of the raw data plays a role in the model training procedure before the model is trained. 
After empirical study and analysis, we observe that each client can only extract a portion of local data that is most relevant to the FL task in the local training.

Our work proposes a new FL framework, \sysname, to achieve the PoLP in FL training for privacy protection. 
% The heart of \sysname is a  newly-proposed local-global dataset distillation.
\sysname can help clients to identify which part of the raw data contributes most to the FL training
and, at the same time, extract the task-relevant information locally.
The central server collects the task-relevant information extracted by clients and distributes the global view to help clients make better information extraction. 
To measure privacy protection effectiveness, we consider adversaries in a semi-honest setting and perform two attacks to infer task-irrelevant information from the data extracted during the training.
% The first attack is a newly-proposed attribute inference attack, which is inspired by the existing secondary inference attack and infer task-irrelevant attributes from the distilled samples.
% The second attack is a membership inference attack.
Experimental results show that \sysname can achieve comparable accuracy to the vanilla FL and better protection for information irrelevant to the training task.
Our contribution is fourfold:
\begin{itemize}
	\item We are the first to introduce the PoLP to the FL training for privacy protection. Data owners can control the local information shared among FL participants and minimize privacy breaches by only sharing the essential FL task-relevant information.
	\item We design a privacy-preserving FL system, \sysname, to realize the PoLP. The key design of \sysname is a local-global dataset distillation, which can identify and extract the task-relevant information gradually by round.
	\item We design a task-irrelevant attribute inference attack to measure the protection effectiveness. The attribute inference attack is inspired by existing secondary inference attacks and fully considers the information leakage in each round.
	\item We implement the system and perform an extensive evaluation. The experimental results show that \sysname can prevent adversaries from inferring task-irrelevant information and preserve high data usability.
\end{itemize}
% \begin{itemize}
% 	\item We are the first to introduce the PoLP to the FL training for privacy protection. Data owners can control the local information shared among FL participants and minimize privacy breaches by only sharing the essential FL task-relevant information.
% 	\item We design a privacy-preserving FL system, \sysname, to realize the PoLP. The key design of \sysname is a local-global dataset distillation, which can identify and extract the task-relevant information gradually by round.
% 	\item We implement the system and perform an extensive evaluation. The experimental results show that \sysname can prevent adversaries from inferring task-irrelevant information and preserve high data usability.
% \end{itemize}

% We promise to open-source all code once the paper is accepted.
% We also plan to launch an online challenge after the paper is accepted and invite the community to design privacy attacks on our distilled data.

% To mitigate the privacy leakage problem in a practical way, some researchers have tried to apply differential privacy \cite{abadi2016deep,dwork2006differential}, \textit{de factor} standard notation for providing rigorous privacy guarantees, to the uploaded information. 

% However, as shown by detailed analysis (Section~\todo{ref}) and empirical study in work~\cite{}, DP-based solutions suffer from deteriorated model performance to a certain extent. 

\section{Related Works}
\label{sec:bg}

%
%The \textit{first} line of privacy-preserving solutions is based on encryption.
%Homomorphic encryption (HE) provides secure protection of uploaded information and training data from being espied by adversaries. 

%The \textit{second} line of solutions is based on a trusted execution environment (TEE).
%Such solutions protect the information sent by the client from being accessed by adversaries through TEE.

\subsection{Privacy-preserving FL}
\label{subsec:ppfl_bg}

Differential privacy (DP) is a typical prior-independent method to offer strong privacy guarantees against adversaries that may infer arbitrary side information~\cite{abadi2016deep}.
LDP, as the variant of the DP, is usually used in FL to mitigate the privacy risk caused by possible gradient leakage by adding noise~\cite{seif2020wireless}. 
DP-based methods are widely-used in FL secnarios~\cite{hu2024communication, shi2024federated, yang2023dynamic}. 
However, these efforts have not done very well in balancing model accuracy and privacy protection.

Some efforts protect the data privacy via information reduction-based solutions.
The goal of these efforts is to make the transformed data as free of private information as possible, thus achieving privacy protection \cite{wu2021dapter,gao2021privacy,xin2020private,wu2024concealing}.
However, these works are primarily applied in DL inference scenarios or for data release. 
How to apply information reduction in FL training scenarios are not explored.
% Work \cite{gao2021privacy} prevents privacy leakage by performing data augmentation on the original data.
% Work \cite{xin2020private} uses the idea of data synthesis to generate a new set of datasets to replace the original data for training.
% Work \cite{wu2024concealing} obfuscates the gradients of the sensitive data with concealed samples. 

\subsection{Dataset Distillation}
\label{subsec:dd}

Dataset distillation (a.k.a. dataset condensation) \cite{yu2023dataset} is a technique to compress a huge dataset to a small one, which can be used to train DL models efficiently. 
% The goal of the dataset distillation is that the model trained on the distilled dataset has comparable accuracy with the one trained on the original dataset. 
\textit{Note that dataset distillation is orthogonal to model knowledge distillation}~\cite{hinton2015distilling}. 
Model distillation aims to transfer the knowledge learned by one model (a.k.a. teacher model) to another model (a.k.a student model).
The model distillation is usually used to perform model compression, where the parameter number of the teacher model is larger than that of the student model.
Many efforts have been expended to continuously improve the usability of dataset distillation~\cite{sucholutsky2021soft,zhao2023dataset}.
Existing dataset distillation approaches are designed for centralized settings to improve training efficiency. Simple solutions, like local distillation without sharing knowledge, fail to produce accurate models. 

\section{Preliminaries}
\label{sec:ow}

\subsection{PoLP in FL}
\label{subsec:polp_in_fl}
Our Principle of Least Privilege (PoLP) idea in FL refers to minimizing the amount of shared information between clients and the central server to include only what is necessary for the primary task. 
In FL, clients share extracted information from their local data to train a global model. 
PoLP aims to ensure that the shared information only contains data relevant to the primary task, denoted as $P$, without exposing sensitive or unnecessary information, which could lead to privacy breaches. 
The goal is to balance effective model training with data privacy by ensuring that shared information $E$ is strictly aligned with the primary task and excludes any irrelevant or sensitive attributes. 
If additional information is shared beyond $P$, it can compromise privacy, and if $E$ does not fully capture $P$, the model's performance may suffer.

\subsection{Security Model}

We assume all participants, i.e., clients and the central server, are semi-honest and follow the \sysname's training protocol. 
All participants faithfully execute the \sysname's training protocol. 
An adversary may be curious about the sensitive data of other participants that are not relevant to the training task, denoted as task-irrelevant information.
The adversary tries to infer the task-irrelevant information from the shared information $E$.

\subsection{Design Challenges}

Bringing PoLP to FL is a challenging task. 
Because 1) the task-relevant information $P$  is not predictable before training, and 2) the task-relevant information $P$ is not element-level separable in raw data $X$. 
In the vanilla FL setting, we cannot control which piece of the information is memorized by local models. 
To control shared information effectively, we must solve two key steps, i.e., identifying information $P$ and extracting information $P$.
% \begin{enumerate}
%      \item Identifying information $P$, i.e., figuring out which part of the information is FL training-relevant.
%      \item Extracting information $P$, i.e., extracting out the identified information $P$ and only sharing this part of the information with other clients.
% \end{enumerate}
% Without loss of generality, we assume that the training dataset contains  $m + 1$ attributes, i.e., $\{P, A_1, A_2, \ldots, A_{m} \}$.
% All task-irrelevant attributes $\{ A_1, A_2, \ldots, A_{m} \}$ may incur privacy leakage.

\section{\sysname's Design}
\label{sec:design}

\begin{figure*}[t]
     \centering
         \includegraphics[width=0.9\textwidth]{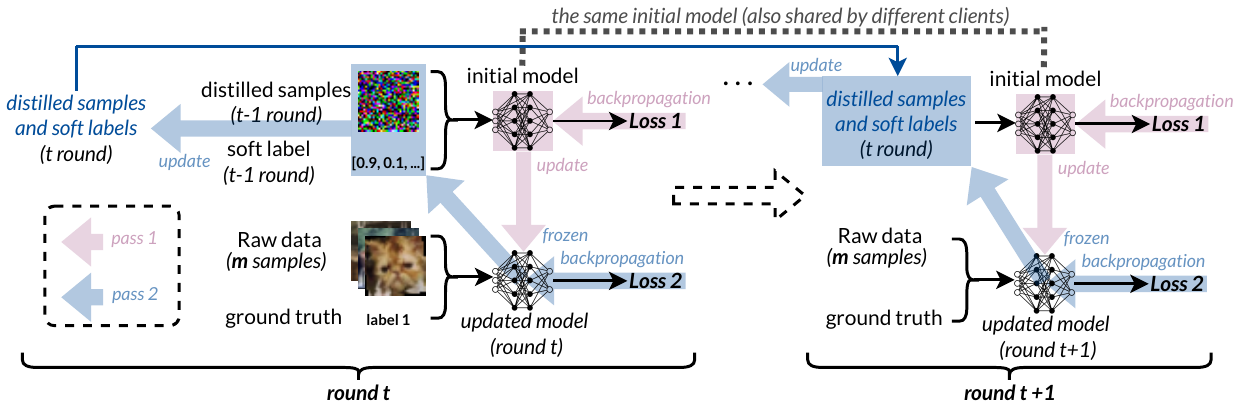}
        \caption{Diagram of the local training algorithm in each client during round t and round t+1. The pass 1 represents the line~\ref{alg_line:update_model},  and the pass 2 represents the lines~\ref{alg_line:update_s}$\sim$\ref{alg_line:update_ita} in Algorithm~\ref{alg:fl_data_train}.}
        \label{fig:local_train_algo}
        % \Description{Diagram of the local training algorithm in each client during round t and round t+1.}
\end{figure*}

% In the following part, we propose a local-global mechanism to address this challenge. 
% In our design,  clients can identify and distill training task-relevant knowledge locally and share the knowledge globally under the coordination of the server to finish the FL.

\subsection{Core Idea}
\label{subsec:pitm}

% Data distillation, as introduced in Section~\ref{subsec:dd}, aims to compress the dataset and maintain the training usability (Equation~\ref{eq:dataset_distillation}).
% It gives us a chance to realize the PoLP in FL.
% We can use a similar idea to extract the training task-relevant information by distilling the raw data samples into a set of distilled samples.
% Then we only share distilled samples with other participants to train the FL model. 
% The privacy leakage can be bounded by the number of distilled samples.
% The smaller the number of distilled samples is, the better the privacy protection effect is.

% Inspired by the idea of dataset distillation, we design a \textit{local-global} dataset distillation method to identify and extract $P$ from the dataset.
% Existing dataset distillation solutions are proposed for the centralized scenario to pursue training efficiency and cannot be deployed in the FL training directly.
% Some simple, straightforward ideas don't work.
% For example, if clients perform distillation locally and do not share task-relevant knowledge among participants, they can not obtain the model with high accuracy.
% And performing distillation centrally on the server does not satisfy the "raw data invisibility" requirement of FL. 

Data distillation, as discussed in Section~\ref{subsec:dd}, compresses the dataset while preserving its utility for training (Equation~\ref{eq:dataset_distillation}). This concept aligns with our Principle of Least Privilege (PoLP), allowing us to extract only the task-relevant information from raw data and share distilled samples with other participants. By limiting the number of distilled samples, we can better protect privacy—the fewer samples shared, the stronger the privacy protection.

Building on this, we propose a \textit{local-global} dataset distillation method to identify and extract key information ($P$) from the data. 
\sysname's training procedure is depicted in Figure~\ref{fig:fl_data}.
Each client needs to extract the training task-relevant information from the raw data and produce a new distilled dataset $\mathcal{S}_i = (S_i, Y_i)$, whose size is much smaller than the size of the original dataset, i.e., $\vert X_i \vert \gg \vert S_i \vert$. 
The distilled dataset $\mathcal{S}_i$ is the information that is shared during training by client $C_i$.
A smaller $\vert S_i \vert$ means a better protection effect.
The model trained on $\mathcal{S}_{\{1,\ldots,n\}}$ can achieve comparable testing performance to that trained on $\mathcal{D}_{\{1,\ldots,n\}}$. That is 
\begin{equation}
\mathbb{E}_{X}[\mathcal{L}(\phi_{\Theta^{\mathcal{D}}}(X), Y)] \simeq	\mathbb{E}_{X}[\mathcal{L}(\phi_{\Theta^{\mathcal{S}}}(X), Y)],
\label{eq:dataset_distillation}
\end{equation}
where $\Theta^{\mathcal{D}}$ and $\Theta^{\mathcal{S}}$ are parameters of the models $\phi_{\Theta^{\mathcal{D}}}$ and $\phi_{\Theta^{\mathcal{S}}}$  trained on dataset $\mathcal{D}$ and $\mathcal{S}$, respectively. 
$\mathcal{L}(\cdot, \cdot)$ is the loss function.

\subsection{\sysname's Design}

We make two assumptions to simplify the description of our design.
1) Each client holds the same kind of categories of samples.
% That is, $\mathsf{card}(Y_i) = \mathsf{card}(Y_i)$, $\forall i,j \in \{1,\ldots,N\}$, where $\mathsf{card}(\cdot)$ is the cardinality of s set.
2) Each client distills $k$ samples per category.
% , and the number of the total distilled samples is $ K = k \times \mathsf{card}(Y_i)$.

\label{subsubsec:core_idea}

%The FL training procedure adopted by \sysname is depicted in Figure~\ref{fig:fl_data}.
%Each client needs to extract the training task-relevant information from the raw data and produce a new distilled dataset $\mathcal{S}_i = (S_i, Y_i)$, whose size is much smaller than the size of the original dataset, i.e., $\vert X_i \vert >> \vert S_i \vert$. 
%The distilled dataset $\mathcal{S}_i$ is the information that is shared during training by client $C_i$.
%A smaller $\vert S_i \vert$ means a better protection effect.
%The model trained on $\mathcal{S}_{\{1,\ldots,n\}}$ can achieve comparable testing performance to that trained on $\mathcal{D}_{\{1,\ldots,n\}}$. That is 
%\begin{equation}
%\mathbb{E}_{X}[\mathcal{L}(\phi_{\Theta^{\mathcal{D}}}(X), Y)] \simeq	%\mathbb{E}_{X}[\mathcal{L}(\phi_{\Theta^{\mathcal{S}}}(X), Y)],
%\label{eq:dataset_distillation}
%\end{equation}
%where $\Theta^{\mathcal{D}}$ and $\Theta^{\mathcal{S}}$ are parameters of the models $\phi_{\Theta^{\mathcal{D}}}$ and $\phi_{\Theta^{\mathcal{S}}}$  trained on dataset $\mathcal{D}$ and $\mathcal{S}$, respectively. 
%$\mathcal{L}(\cdot, \cdot)$ is the loss function.

%To clearly demonstrate the core idea, we make two assumptions here to simplify the description.
%1) Each client holds the same kind of categories of samples. That is, $\mathsf{card}(Y_i) = \mathsf{card}(Y_i)$, $\forall i,j \in \{1,\ldots,N\}$, where $\mathsf{card}(\cdot)$ is the cardinality of s set.
%2) Each client distills $k$ samples per category, and the number of the total distilled samples is $ K = k \times \mathsf{card}(Y_i)$.

\subsubsection{Local Procedure} 

In the local procedure, \sysname extracts the local task-relevant knowledge.
During the initialization, the central server distributes \textit{the same} randomly initialized model $\phi_{\theta_\mathsf{init}}$ to all clients.
And each client randomly initializes the distilled dataset $\mathcal{S}_i$.
When performing the local dataset distillation in round $t$,
each client updates the distilled data through
\begin{equation}
S_i^{t+1}, \tilde{Y}_i^{t+1}, \overrightarrow{\eta}_i^{t+1} = \mathop{\arg\min}_{S_i^t, Y_i^t, \overrightarrow{\eta}_i^t}  \mathcal{L} (\phi_{\theta_i^t}(X_i), Y_i ), 
\label{eq:distill_base}
\end{equation}
where the $\theta_i^t$ is optimized based on  $\theta_\mathsf{init}$ with $K$ distilled samples through
%\begin{equation}
%	\theta_i^t = \theta_\mathsf{init} - \overrightarrow{\eta}_i^t \nabla_{\theta_\mathsf{init}} \mathcal{L}(\theta_\mathsf{init}(S_i^t), \tilde{Y}_i^t),
%\end{equation}
\begin{equation} \Big\{
\begin{aligned}
&~\theta_{i, j+1}^t =  \theta_{i,j}^t - \overrightarrow{\eta}_{i,j}^t \nabla_{\theta_{i,j}^t} \mathcal{L}( \phi_{\theta_{i,j}^t}(S_{i,j}^t), \tilde{Y}_{i,j}^t) \\
&~\theta_{i,0}^t = \theta_\mathsf{init} \\
 \end{aligned},
\end{equation}
where $j$ indicates the $j$-th element of the corresponding  data, and $j \in \{0, \ldots, K-1\}$. 
 $S_{i,j}^t$ is the $j$-th distilled sample extracted from the raw data of client $i$ in the round $t$.
$\tilde{Y}_{i,j}^t$ is the soft label of the distilled data $S_{i,j}^t$. The soft label design, inspired by Work~\cite{sucholutsky2021soft}, can improve the training accuracy by introducing more information. 
$\overrightarrow{\eta}_{i,j}^t$ is the $j$-th sample's learning rate in the round $t$, which is used to optimize the $\theta_i^t$.  
In our design, each distilled sample has a learnable learning rate to maximize the training accuracy of model $\phi_{\theta_i}$.

We train a learning rate for each distilled sample due to the following reasons. 
To ensure the stability of model optimization through data aggregation,
in each round, the $\theta_i^t$ is optimized from the $\theta_\mathsf{init}$, which is initially distributed by the central server. 
However, updating the parameters from the same initial parameters, instead of updating the parameters from the parameters obtained in the last round, affects the training accuracy. 
Therefore, we design a dynamic adjustment mechanism of the learning rate $\overrightarrow{\eta}_i$. 
% All details of the training procedure are introduced in Section~\ref{subsec:desgin_details} through Algorithm~\ref{alg:fl_data_train}.

\subsubsection{Global Procedure} 

In the global procedure, \sysname shares the training-related information extracted locally among all clients. 
% The aggregation methods should 1) aggregate distilled information of all clients, and 2) be compatible with the next round of local distillation.
% Following the idea of parameter aggregation in the vanilla FL setting, we design the following aggregation method.
The central server aggregates the data shared by all clients.
Recall that each client has $K$ locally-distilled samples, soft labels, and learning rates.
The server produces $K$ aggregated samples $\{s_1^*, \ldots, s_K^* \}$, soft labels $\{\tilde{y}_1^*, \ldots, \tilde{y}_K^* \}$, and learning rate $\{\eta_1^*, \ldots, \eta_K^* \}$.  
\begin{equation}
%\begin{aligned}
 s_j^*, \tilde{y}_j^*, \overrightarrow{\eta}_j^*  = \frac{1}{N} \sum_{i=1}^{N} S_{i,j}^t, \frac{1}{N} \sum_{i=1}^{N} \tilde{Y}_{i,j}^t, \frac{1}{N} \sum_{i=1}^{N} \overrightarrow{\eta}_{i,j}^t,
% \end{aligned},
\end{equation}
$S_{i,j}^t$,  $\tilde{Y}_{i,j}^t$, and $\overrightarrow{\eta}_{i,j}^t$ is the $j$-th element of $S_i^t$, $\tilde{Y}_i^t$, and $\overrightarrow{\eta}_i^t$, respectively.
%The distilled samples should be quantitatively aligned by labels, 
%and samples should be sorted in a predetermined order.
Then the server distributes the aggregated samples to clients. 
Note that \textit{\textbf{no} model training is needed on the server, and the model training is on the client side}.

Clients update the distilled local data through Equation~\ref{eq:distill_base} continually.
After a sufficient number of local-global information sharing between the server and clients,  
all clients obtain the FL model by optimizing the $\theta_\mathsf{init}$  \textit{\textbf{with $K$ distilled samples in one epoch locally}.}
The training procedure of the FL model is 
%\begin{equation}
%	\theta_i = \theta_\mathsf{init} - \overrightarrow{\eta}_i \nabla_{\theta_\mathsf{init}} \mathcal{L}(\theta_\mathsf{init}(S_i), \tilde{Y}_i),
%%	\label{eq:fl_data_obj_bases}
%\end{equation}
\begin{equation} \Big\{
\begin{aligned}
&~\theta_{i, j+1} = \theta_{i, j} - \overrightarrow{\eta}_{i,j} \nabla_{\theta_{i, j}} \mathcal{L}(\phi_{\theta_{i, j}}(S_{i,j}), \tilde{Y}_{i,j})\\
&~\theta_{i, 0} = \theta_\mathsf{init}
 \end{aligned},
\end{equation}
where $j$ indicates the $j$-th element of the corresponding data, and $j \in \{0, \ldots, K-1\}$. 
$S_i$, $\tilde{Y}_i$, and $\overrightarrow{\eta}_i$ are the distilled samples, soft labels, and learning rates of client $C_i$ produced in the last round.

% The way used by \sysname to share the local information is different from that of the vanilla FL, where clients share information in the form of local models measured in Equation~\ref{eq:typ_fl_mi}. 
% In \sysname's setting,  clients share their local information through distilled local data, i.e., 
% \begin{equation}
% 	E = \mathcal{I} (\mathcal{S}^*; X),
% 	\label{eq:data_fl_mi}
% \end{equation}
% where $\mathcal{S}^*$ is the set of all distilled samples, soft labels, and learning rates.
% Recall that the smaller the $K$ (i.e., the number of the distilled samples) is, the better the privacy protection is (Section~\ref{subsec:pitm}).

\begin{algorithm}[t]
    \caption{The training algorithm of \sysname's local procedure. \label{alg:fl_data_train}}
    \label{alg:algorithm}
    \textbf{Input}: Number of the distilled samples: $K$; 
    Epoch number of local distillation: $\mathsf{Eps}$;
    Initial model parameter: $\theta_\mathsf{init}$; 
    Local raw data: $\mathcal{D}=\{(x_j, y_j)\}_{j\in\{1,\ldots, \vert \mathcal{D} \vert \}}$ \\
    \textbf{Parameter}: The learning rate for optimizing the distilled data: $\gamma$; 
    Aggregated samples after $t$ rounds:  $S^t = \{s_j^t\}_{j\in\{1,\ldots, K \}}$;
    Aggregated soft labels after $t$ rounds: $\tilde{Y}^t = \{\tilde{y}_j^t\}_{j\in\{1,\ldots, K \}}$;
    Aggregated learning rate after $t$ rounds: $\overrightarrow{\eta}^t = \{ \eta_j^t \}_{j\in\{1,\ldots, K \}}$\\
    \textbf{Output}: Distilled data of $(t+1)$-th round : $S^{t+1}, \tilde{Y}^{t+1}, \overrightarrow{\eta}^{t+1}$
    \begin{algorithmic}[1] %[1] enables line numbers   
        \STATE $ S_0^{t}, \tilde{Y}_0^{t}, \overrightarrow{\eta}_0^{t} = S^{t}, \tilde{Y}^{t}, \overrightarrow{\eta}^{t}$
        %\Comment{initialization}
         	\FOR {$e$  in $\{1, 2, \dots, \mathsf{Eps}\}$}
                    \FOR {$\{(x_\mathsf{bh},y_\mathsf{bh})\}$ in $\mathcal{D}$}  
                    \label{alg_line:batch}
                        \FOR {$k$ in $\{1, \dots, K\}$}
                        \label{alg_line:loop_update}
                        \IF{$k == 1$}
                            \STATE $\theta_\mathsf{tmp} = \theta_\mathsf{init}$\;
                        \ELSE
                            \STATE $\theta_\mathsf{tmp} = \theta_{k-1}^t$\;
						\ENDIF
                        \STATE $\theta_{k}^t = \theta_\mathsf{tmp} - \eta_k^t \nabla_{\theta_\mathsf{tmp}}(\phi_{\theta_\mathsf{tmp}}(s_k^t), \tilde{y}_k^t) $ \;\label{alg_line:update_model}
                        \ENDFOR
                        \STATE $L_\mathsf{raw} = \mathcal{L}(\phi_{\theta_K^t}(x_\mathsf{bh}), y_\mathsf{bh})$\;\label{alg_line:cal_loss}
                        \STATE $S_{e-1}^t = S_{e-1}^t - \gamma \nabla_{S_{e-1}^t} L_\mathsf{raw}$ \; \label{alg_line:update_s}
         			\STATE $\tilde{Y}_{e-1}^t = \tilde{Y}_{e-1}^t - \gamma \nabla_{\tilde{Y}_{e-1}^t} L_\mathsf{raw}$ \;
         			\STATE $\overrightarrow{\eta}_{e-1}^t = \overrightarrow{\eta}_{e-1}^t - \gamma \nabla_{\overrightarrow{\eta}_{e-1}^t} L_\mathsf{raw}$ \;\label{alg_line:update_y}
                        \label{alg_line:update_distilled_data} \label{alg_line:update_ita}
                    \ENDFOR
                    \STATE $S_{e}^{t+1}, \tilde{Y}_{e}^{t+1}, \overrightarrow{\eta}_{e}^{t+1} = S_{e-1}^{t}, \tilde{Y}_{e-1}^{t}, \overrightarrow{\eta}_{e-1}^{t}$\;
                \ENDFOR
                \STATE $S^{t+1}, \tilde{Y}^{t+1}, \overrightarrow{\eta}^{t+1} = S_{\mathsf{Eps}}^{t}, \tilde{Y}_{\mathsf{Eps}}^{t}, \overrightarrow{\eta}_{\mathsf{Eps}}^{t}$\;
        \STATE \textbf{return} $S^{t+1}, \tilde{Y}^{t+1}, \overrightarrow{\eta}^{t+1}$ \label{alg_line:return}
    \end{algorithmic}
\end{algorithm}

\subsection{Local Training Details}
\label{subsec:desgin_details}

In this part, we report the design details of the local procedure through Algorithm~\ref{alg:fl_data_train}.
Recall that the local training procedure is performed on the client side.
In the first round, the central server initializes $K$ distilled samples and the corresponding soft labels and learning rates for all clients. 
The server also initializes the model with parameter $\theta_\mathsf{init}$. 
During the local procedure, the client extracts the training task-relevant information from the distilled data by batch (line~\ref{alg_line:batch}). 
\sysname uses the aggregated data in round $t$ to optimize the parameter $\theta$ of local model by sample (line~\ref{alg_line:update_model}).
Once the parameter is optimized with $K$ samples, \sysname calculates the loss of the updated model with the raw data (line~\ref{alg_line:cal_loss}). 
And the loss will be used to update the distilled samples, soft labels, and learning rates (line~\ref{alg_line:update_distilled_data}).
After $\mathsf{Eps}$ distillation epochs, the client obtains the distilled data of $(i+1)$-th round (line~\ref{alg_line:return}). 
We demonstrate the algorithm details in Figure~\ref{fig:local_train_algo}, the pass 1 represents the line~\ref{alg_line:update_model} and the pass 2 represents the lines~\ref{alg_line:update_s}$\sim$\ref{alg_line:update_ita}. 
Note that the initial model is randomly initialized by the central server and shared by all clients.

% For typographical purposes
\begin{table*}[t]
\centering
\caption{The top-1 accuracy of the vanilla FL and our \sysname (reported in \%). V stands for vanilla FL; $F_k$ stands for each category containing $k$ distilled samples in \sysname's settings.}
\label{tab:accuracy_ass}
\begin{tabular}{rcccccccccccc}
\toprule
                  & \multicolumn{4}{c}{\textbf{TinyResNet}}                                                                   & \multicolumn{4}{c}{\textbf{AlexNet}}                                                                 & \multicolumn{4}{c}{\textbf{ConvNet}}                                                                  \\ \cmidrule(lr){2-5} \cmidrule(lr){6-9} \cmidrule(lr){10-13}
                  & V     & $F_{10}$ & $F_{15}$ & $F_{20}$ & V     & $F_{10}$ & $F_{15}$ & $F_{20}$ & V     & $F_{10}$ & $F_{15}$ & $F_{20}$ \\ 
\midrule
\textbf{MNIST}    & 99.03 & 95.56    & 95.51    & 95.17    & 99.26 & 95.10    & 96.33    & 96.05    & 99.32 & 99.06    & 99.19    & 99.19    \\ 
\textbf{CIFAR-10} & 79.88 & 68.73    & 68.91    & 70.09    & 72.46 & 67.77    & 70.48    & 71.27    & 86.76 & 77.18    & 81.19    & 81.79    \\ 
\textbf{CIFAR-100}& 44.33 & 44.04    & 44.50    & 44.63    & 41.17 & 43.21    & 46.13    & 48.86    & 58.85 & 48.71    & 56.07    & 53.15    \\ 
%\textbf{SVHN}     & 91.68 & 78.01    & 77.74    & 78.25    & 91.75 & 72.00    & 82.06    & 83.91    & 93.73 & 90.02    & 92.25    & 91.86    \\ 
\bottomrule
\end{tabular}
\end{table*}

\section{Privacy Assessment Method}
\label{sec:tm}

% In the light of the PoLP introduced in Section~\ref{subsec:polp_in_fl}, the privacy leakage risks come from the mutual information between the distilled information and the sensitive attributes. 
% That is $\mathcal{I}(E;Z)$, where $E$ is calculated through Equation~\ref{eq:data_fl_mi}, and $Z$ is the sensitive information of the raw data.  
% We protect privacy by removing training task-irrelevant information from the shared information $E$ (introduced in Section~\ref{subsec:pitm}). 

% There is no privacy attack that can be directly performed on the locally distilled data.
We design the following two attacks to evaluate the privacy protection ability of our \sysname. 

\subsection{Task-irrelevant Attributes Inference Attack}

In the proposed attack, a passive adversary tries to infer the task-irrelevant attributes of the raw data from the distilled data.
To mimic the adversary, 
we first train a binary classification model to distinguish the samples belonging to category $l_{p^1}$ from a set of samples belonging to categories $l_{p^1}$ and $l_{s^1}$. 
Note that each sample has and only has one label, i.e., $l_{p^1}$ or not $l_{p^1}$. 
We denote the binary classification model as $\psi_{p^1-s^1}$, and we say that the $l_{p^1}$ is the primary attribute and the $l_{s^1}$ is the task-irrelevant attribute.
The classifier can perform the following classification:
\begin{equation}
\psi_{p^1-s^1}(X) = 
\begin{cases}
1, \text{if}  ~ X ~\text{belongs to} ~ l_{p^1} \\
0, \text{otherwise}
\end{cases}	
\end{equation}
Similarly,  we train another three binary classification models, i.e., $\psi_{p^2-s^2}$, $\psi_{p^1-s^2}$, and $\psi_{p^2-s^1}$, where $l_{p^2}$ and $l_{s^2}$ are another two categories different from $l_{p^1}$ or $l_{s^1}$.
According to the protection goals, the distilled data in all four tasks should only contain primary attributes, i.e., $l_{p^1}$ or $l_{p^2}$ and not contain task-irrelevant attributes, i.e., $l_{s^1}$ or $l_{s^2}$.

Then, we train a binary classification model $\psi_{s^1-s^2}$ with the distilled data collected during the training procedure of models $\psi_{p^1-s^1}$ and $\psi_{p^1-s^2}$.
Next, we use the trained model $\psi_{s^1-s^2}$ to classify the distilled data collected in the training procedure of $\psi_{p^2-s^1}$ and $\psi_{p^2-s^2}$.
The classification accuracy (denoted as attack accuracy) can indicate the protection effectiveness.
If our \sysname works as expected, the testing accuracy should be about 50\%, because both test data and training data hardly contain the features of the $l_{s^1}$ and $l_{s^2}$ categories.

\subsection{Membership Inference Attack}

In this attack, adversaries aim to determine whether a particular sample is included in the training data.
Here we denote the membership inference attack as $\mathcal{A}_\mathsf{mia}(x, \theta) \in [0,1]$, where the $\mathcal{A}_\mathsf{mia}(\cdot)$ is the target function to infer whether the input $x$ comes from the parameter $\theta$'s training data. 
If $\mathcal{A}_\mathsf{mia}(x, \theta) = 1$, the input $x$ belongs to the training data, not vice-versa.
The corresponding optimization goal is:
\begin{equation}
    \min_{\mathcal{A}_\mathsf{mia}} \mathbb{E}_{x\in X}[\mathcal{L}(\mathcal{A}_\mathsf{mia}(x, \theta),m(x))],
\end{equation}
where $X$ is the collection of distilled samples in all rounds.
$m(\cdot)$ is the sensitive member indicator, and 
\begin{equation}
    m(x) = \begin{cases}
    1 & \text{If } x \text{ comes from the training dataset}\\
    0 & \text{Otherwise}
    \end{cases} .
\end{equation}
%$m(\cdot)=1$, if the sample belongs to the training data, and $m(\cdot)=0$, if the sample does not belong to the training data. 
$\mathcal{L}$ is the loss function, e.g., Mean Square Error function in our context.

To construct this attack, we train several shadow models $$\phi_{\psi_1}, \phi_{\psi_2}, \cdots, \phi_{\psi_M}$$ to imitate the behavior of the model to attack, i.e., $\phi_\theta$. 
We split the original training dataset and test dataset into several sub-datasets for training the corresponding shadow model. 
Then, we use the outputs of shadow models to train a binary classifier to achieve the attack through the optimization objective:
\begin{equation}
    \min_{\mathcal{A}_\mathsf{mia}} \frac{1}{M}\sum_{j=1}^{M}\mathbb{E}_{x\in X_\mathsf{train} \cup X_\mathsf{test}}[\mathcal{L}(\mathcal{A}_\mathsf{mia}(x, \psi_j),m(x))]
\end{equation}

\section{Evaluation}
\label{sec:eval}

\begin{figure*}[t]
     \centering
         \includegraphics[width=\textwidth]{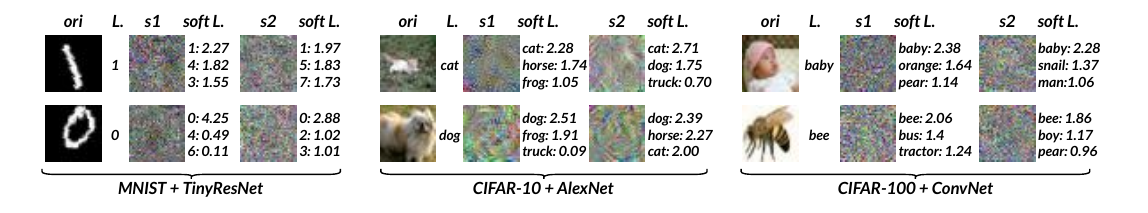}
        \caption{Examples of the distilled samples and soft labels in different settings. (soft) L. stands for the (soft) label. The s1 and s2 denote the distilled samples. We report the top 3 elements of the corresponding soft label with confidence.}
        \label{fig:samples}
%        \vspace{-5pt}
        % \Description{Examples of the distilled samples and soft labels in different settings.}
\end{figure*}

\begin{table*}[t]
\centering
\caption{The attack accuracy of the task-irrelevant attributes inference (reported in \%). Accuracy around 50\% is equivalent to a random guess.}
\label{tab:secondary_inference}
\begin{tabular}{llccccccccc}
\toprule
ID & Dataset & $l_{p^1}$ & $l_{p^2}$ & $l_{s^1}$ & $l_{s^2}$ & $\psi_{p^1-s^1}$ Acc & $\psi_{p^1-s^2}$ Acc & $\psi_{p^2-s^1}$ Acc & $\psi_{p^2-s^2}$ Acc & \textbf{Attack Acc} \\
\midrule
1  & \multirow{6}{*}{CIFAR-10} & horse       & frog        & automobile  & cat         & 95.05                 & 84.70                 & 95.65                 & 83.65                 & \textbf{46.88}      \\
2  &                            & horse       & airplane    & deer        & dog         & 79.80                 & 84.25                 & 91.50                 & 93.80                 & \textbf{48.38}      \\
3  &                            & airplane    & horse       & cat         & automobile  & 90.85                 & 90.70                 & 84.70                 & 95.05                 & \textbf{49.94}      \\
4  &                            & frog        & horse       & automobile  & deer        & 95.65                 & 84.90                 & 95.05                 & 79.80                 & \textbf{50.00}      \\
5  &                            & frog        & airplane    & bird        & dog         & 84.20                 & 88.80                 & 86.65                 & 93.80                 & \textbf{50.00}      \\
6  &                            & airplane    & frog        & cat         & automobile  & 90.85                 & 90.70                 & 83.65                 & 95.65                 & \textbf{54.69}      \\
\midrule
7  & \multirow{6}{*}{CIFAR-100} & bowl        & chair       & bear        & bed         & 90.50                 & 86.50                 & 96.50                 & 88.00                 & \textbf{50.00}      \\
8  &                            & bowl        & beetle      & bear        & bed         & 90.50                 & 86.50                 & 88.00                 & 92.50                 & \textbf{50.00}      \\
9  &                            & bowl        & chair       & baby        & bear        & 87.00                 & 90.50                 & 93.00                 & 96.50                 & \textbf{51.13}      \\
10 &                            & chair       & bowl        & baby        & bed         & 93.00                 & 88.00                 & 87.00                 & 86.50                 & \textbf{58.31}      \\
11 &                            & beetle      & bowl        & baby        & bear        & 94.50                 & 92.50                 & 87.00                 & 90.50                 & \textbf{58.88}      \\
12 &                            & beetle      & chair       & baby        & bed         & 94.50                 & 92.50                 & 93.00                 & 96.50                 & \textbf{63.37}      \\
\bottomrule
\end{tabular}
\end{table*}

\subsection{Implementation}

We perform the evaluation on a server with two A100 (80G) GPU cards.
We use three datasets, i.e.,  MNIST~\cite{lecun1998gradient}, CIFAR-10~\cite{krizhevsky2009learning}, and CIFAR-100~\cite{krizhevsky2009learning}, and three model architectures, i.e., TinyResNet, AlexNet~\cite{krizhevsky2017imagenet}, and ConvNet~\cite{zhao2021dataset}.
The TinyResNet consists of a convolution layer, whose weight shape is $3\times7\times7\times64$, and a ResBlock~\cite{he2016deep}, whose kernel size is $3\times3$ and output channel is 64.
% Tiny-ResNet, three convolution layers, one pooling layer, and one linear layer:
% \begin{enumerate}
%     \item 1-th conv: kernel size=7, stride=3, out\_channel=64
%     \item maxpool 2d, kernel size=3, stirde=3
%     \item 2-th conv: kernel size=3, stride=1, out\_channel=64, padding=1
%     \item 3-th conv: kernel size=3, stride=1, out\_channel=64, padding=1
%     \item linear: 1024 x 10
% \end{enumerate}
% there is a short cut from 2 to 4.
%$\eta_i$ in Equation~\ref{eq:typical_fl_update} and
To train models in the vanilla FL settings and in our setting, we use one central server and five clients.
We initialize the learning rate to 0.01 and gradually decrease as the training progresses ($\gamma$ in Algorithm~\ref{alg:fl_data_train}).  
The batch size to distill the samples is 1024 (line~\ref{alg_line:batch} in Algorithm~\ref{alg:fl_data_train})\footnote{When training the ConvNet on CIFAR-100, we initialize the learning rate $\gamma$ with 0.005 and set the batch size to 512.}.
The learning rate $\gamma$ is multiplied by a scaling factor of 0.5 after every 40 aggregation rounds, and the overall number of aggregation rounds is set to 100.
For \sysname's training, we set the epoch number ($\mathsf{Eps}$ in Algorithm~\ref{alg:fl_data_train}) to 30.
For the vanilla FL training, we use the FedAvg method to aggregate the local models.

%\begin{table}[t]
%\begin{tabular}{c|c|c|c|c}
%\hline
%          & \textbf{MNIST} & \textbf{CIFAR-10} & \textbf{CIFAR-100} & \textbf{SVHN} \\ \hline
%T         &                &                   &                    &               \\ \hline
%$F_{10}$ &   1.1768 MB             &                   &        12.1117 MB            &               \\ \hline
%$F_{15}$ &          1.7653 MB      &                   &       18.1675 MB             &               \\ \hline
%$F_{20}$ &        2.3537 MB        &                   &         24.2233 MB           &               \\ \hline
%\end{tabular}
%\caption{The size of the model trained in typical settings, and the size of distilled samples in \sysname settings.}
%\label{tab:statistics}
%\end{table}

\subsection{Accuracy Assessment}

We report the top-1 accuracy of \sysname and vanilla FL by dataset and model architecture in Table~\ref{tab:accuracy_ass}.
In these experiments, we randomly divide the training dataset equally among five clients.
In the table, V stands for the model trained in the vanilla FL setting. 
$F_k$ stands for the model trained by \sysname, where the subscript $k$ indicates the number of distilled samples per category. 
For example, the dataset CIFAR-10 consists of 10 categories. Therefore, $F_{10}$ indicates the model trained with 100 distilled samples, i.e., $K$=100.  
Recall that the smaller the $k$ is, the better the protection effect is.
We have the following observations based on the experimental results.  

\textit{First}, our \sysname achieves an accuracy close to that of the vanilla method in the majority of all 27 experiments.
And the accuracy of \sysname exceeds that of the vanilla FL in 5 experiments. 
In the rest of the experiments, the accuracy of \sysname lags behind that of the vanilla FL by only 3\% in 8 experiments and by 5\% in 15 experiments.

\textit{Second},
the accuracy of \sysname on the dataset CIFAR-100 is comparable to that of vanilla FL and even performs better in some configurations, i.e., TinyResNet-$F_{15/20}$ and AlexNet-$F_{10/15/20}$.
In terms of relative accuracy to the vanilla method, our \sysname performs better on CIFAR-100 than that on CIFAR-10 and MNIST.
This is because the number of distilled samples per category in CIFAR-100 (i.e., 500 samples) is less than that of the other two datasets (i.e., about 5k$\sim$6k samples). 
So, the distilled samples in $F_{20}$ setting contain more information about the CIFAR-100 dataset compared to that of the other two datasets.

\textit{Third}, 
as the number of distilled samples, i.e., $K$, increases, the model's accuracy is improved under all experimental configurations. 
For every 5 additional distilled samples per category, the accuracy increases by an average of 0.12\%, 2.172\%, and 3.985\% on MNIST, CIFAR-10, and CIFAR-100, respectively.
There are few abnormal cases where a larger number of distilled samples does not gain better accuracy. 
For example, the accuracy of ConvNet trained on CIFAR-100 with $k=20$ is less than that trained with $k=15$.
We believe that the outlier does not affect the overall conclusions, considering the highly nonlinear nature of the DL model.

Please note that we will set the value of $k$ to 20 in the following privacy assessment, because under this configuration, \sysname has higher accuracy and higher risk of privacy leakage.
%We report the communication efficiency of both the typical FL and \sysname in Table~\ref{tab:statistics}.
%That is the size of the model trained in typical settings and the size of distilled samples in \sysname settings.
%Note that the statistical information only shows that \sysname achieves higher accuracy with a smaller communication cost than the traditional approach.
%For example, in the experimental configuration of ConvNet and SVHN, \sysname's accuracy ($F_{20}$) is \todo{XX} higher than that of traditional FL, but the communication size is XXX. 

\subsection{Aggregated Samples Visualization}
\label{subsec:asv}

Given that distilled samples in our experiments are in the format of images, we show some distilled samples and the corresponding soft labels in Figure~\ref{fig:samples}.
We present one original sample and two aggregated samples in the corresponding category in the $F_{20}$ configuration.
We also report the categories and the corresponding scores of the top three highest elements in the soft labels (ranked by the score).
We can see that the distilled samples shared among participants are \textit{visually obfuscated}.
Distilled samples of different datasets in different categories all seem randomly generated. 
In the next part, we will perform the privacy assessment on the distilled samples during the entire training.

%four datasets X (raw data, raw label ->  aggregated data, aggregated label) X 2 category.

\subsection{Privacy Assessment}
\label{subsec:pri_ass}

% In our context, we try to remove as much as possible task-irrelevant information from the training dataset. 
% The task-relevant information is maintained, which is shown by the accuracy measurement. 
% Here we perform privacy assessment through the attack introduced in Section~\ref{sec:tm}.

\subsubsection{Task-irrelevant Attribute Inference}
To perform the measurement, we select the categories $l_{p^1}$, $l_{p^2}$, $l_{s^1}$, and $l_{s^2}$ from the dataset CIFAR-10 and CIFAR-100, and we use TinyResNet as train the following binary classification models $\psi$. 
We distill 20 samples per category, which is the same setting as the $F_{20}$ configuration used in the accuracy assessment (Table~\ref{tab:accuracy_ass}).

We perform 12 attacks and report the attack accuracy in Table~\ref{tab:secondary_inference}. 
Note that if the attack accuracy around 0.5 is equivalent to a random guess in the $\psi_{s^1-s^2}$ classification.
We find that 7 of 12 attacks can effectively defend against task-irrelevant inference attacks, where the attack accuracy equals to or is less than 0.5. 

We find that the protection effect on CIFAR-100 is worse than that on CIFAR-10, e.g., Attack 10$\sim$12.
This is because, for CIFAR-10, we distill 20 samples from 5,000 original samples per category, and for CIFAR-100, the 20 samples per category are distilled from 500 original samples. 
So the samples distilled from CIFAR-100 contain more information about the original samples than that of CIFAR-10.
However, compared to primary classification tasks with an average accuracy above 90\%, 
the results of Attack 10$\sim$12 still indicate that our \sysname can greatly reduce the task-irrelevant attributes and have a good privacy protection ability.

% \begin{table}[t]
% \centering
% \caption{Results of the membership inference attack (reported in \%). V denotes vanilla FL, and $F_{20}$ is our \sysname with $k=20$. The model architecture is AlexNet, and the training dataset is CIFAR-10.}
% \label{tab:mia_results}
% \begin{tabular}{lcc}
% \toprule
% Epoch    & V       & $F_{20}$  \\ 
% \midrule
% \textbf{1}   & 50.09 & 49.75 \\ 
% \textbf{25}  & 51.51 & 49.75 \\ 
% \textbf{50}  & 52.95 & 49.75 \\ 
% \textbf{75}  & 57.12 & 49.75 \\ 
% \textbf{100} & 59.14 & 49.75 \\ 
% \bottomrule
% \end{tabular}
% \end{table}

\subsubsection{Membership Inference Attack}

We perform the membership inference attack with AlexNet and CIFAR-10. 
$k$ values of  our \sysname is 20. 
Setting the value of $k$ to 20, the data distilled by the \sysname contains more information about the training dataset and thus the risk of privacy leakage is higher.
In each attack, we train 128 shadow models.
We use CatBoost as the binary classifier.
We set the depth to 1 and the number of iterations to 10000.
We perform the attack on models produced at round 1, round 25, round 50, round 75, and round 100, respectively.
% , and report the results in Table~\ref{tab:mia_results}.
The experimental results show that the vanilla FL is vulnerable to membership inference attack, and the attack accuracy is 59.14\%, while our method can resist this attack with an attack accuracy of only 49.75\%.
That is, the adversaries cannot determine whether the given samples exist in the training datasets, and \sysname   can effectively defend against membership inference attacks.

\section{Conclusion}

We design a new FL system, \sysname, that brings the Principle of Least Privilege (PoLP) to federated learning (FL) scenarios by extracting the FL task-relevant information lying in the raw data. 
\sysname gives users a chance to control the amount of share information during the FL training and make a better trade-off between usability and privacy protection.
Experimental results show that \sysname can effectively protect the training task-irrelevant information and maintain the training accuracy in the semi-honest setting. 
We hope our \sysname could attract more research efforts in achieving controllable privacy protection, or more broadly, the data right problems in the scenario of FL training.

\bibliographystyle{IEEEtran}
\bibliography{ref}

\end{document}